\documentclass[article,10pt]{IEEEtran}
\usepackage{cite}

\ifCLASSINFOpdf
  \usepackage[pdftex]{graphicx}
  \usepackage{epstopdf}
  \AppendGraphicsExtensions{.tif}
  
  \DeclareGraphicsExtensions{.eps}
\else
\fi
\usepackage[cmex10]{amsmath}
\usepackage{array}

\usepackage{mdwmath}
\usepackage{mdwtab}
\usepackage[caption=false,font=footnotesize]{subfig}
\usepackage{fixltx2e}
\usepackage{hhline}
\usepackage{tikz}
\usepackage{pgfplots}
\usepackage{epstopdf}
\usepackage{dblfloatfix}
\usepackage{wrapfig}
\usetikzlibrary{arrows}
\usetikzlibrary{decorations.pathreplacing}
\usetikzlibrary{patterns}
\usetikzlibrary{circuits.logic.US,circuits.logic.IEC}

\usepackage{xcolor}

\begin{document}
\bstctlcite{IEEEexample:BSTcontrol}
%
\title{\huge Minimum Energy Quantized Neural Networks}

\author{
\IEEEauthorblockN{Bert Moons$^+$, Koen Goetschalckx$^+$, Nick Van Berckelaer* and Marian Verhelst$^+$\\}
\IEEEauthorblockA{Department of Electrical Engineering*$^+$ - ESAT/MICAS$^+$, KU Leuven, Leuven, Belgium}
\vspace{-3ex}
}


%


\maketitle
\begin{abstract}
This work targets the automated minimum-energy optimization of Quantized Neural Networks (QNNs) - networks using low precision weights and activations. These networks are trained from scratch at an arbitrary fixed point precision. At iso-accuracy, QNNs using fewer bits require deeper and wider network architectures than networks using higher precision operators, while they require less complex arithmetic and less bits per weights. This fundamental trade-off is analyzed and quantified to find the minimum energy QNN for any benchmark and hence optimize energy-efficiency.
To this end, the energy consumption of inference is modeled for a generic hardware platform. This allows drawing several conclusions across different benchmarks. First, energy consumption varies orders of magnitude at iso-accuracy depending on the number of bits used in the QNN. Second, in a typical system, BinaryNets or int4 implementations lead to the minimum energy solution, outperforming int8 networks up to $2-10\times$ at iso-accuracy. All code used for QNN training is available from https://github.com/BertMoons/.
\end{abstract}


%
\IEEEpeerreviewmaketitle

\section{Introduction}
\label{sec:introduction}
Deep learning~\cite{lecun2015deep}, and more specifically Convolutional Neural Networks (CNNs) have come up as state-of-the-art classification algorithms, achieving super-human performance in applications in both computer vision (CV) and automatic speech recognition. 
Although these networks are extremely powerful, they are also very computationally and memory intensive, making them difficult to employ on embedded or battery-constrained systems. Today, training, or even neural network inference is therefore run on specialized, very fast and power hungry Graphical Processing Units (GPU). Substantial research efforts are spent in either speeding up or minimizing energy consumption of NNs at run-time on both general-purpose and specialized computer hardware. 

Minimizing energy consumption is especially crucial in Neural Networks used in battery constrained, wearable and always-on applications. In these systems, inference at the edge is crucial, in order to reduce the latency and wireless connectivity costs as well as the privacy concerns associated with cloud-connectivity. Previous solutions where insufficient for such purposes, as both the used hardware platform and the used Neural Networks are not sufficiently energy-efficient for always-on, low-latency processing. A simple always-on face-detection task on this platform drains its battery in less than 40 minutes~\cite{likamwa2014draining}. In this work, we propose a framework to analyze and optimize the trade-off between energy-consumption and accuracy for Quantized Neural Networks (QNNs), hence allowing co-design of hardware and algorithm towards minimum energy consumption in any application.

\section{Related Work}
\label{sec:related_work}
Several approaches have been proposed to reduce the energy footprint of DNNs. Most efforts are either in designing more efficient algorithms, or in designing optimized hardware.

\textbf{Dedicated hardware platforms} are optimized for typical dataflows and exploit network sparsity as well as the inherent error-resilience of most Neural Networks. The highly parallel nature of DNNs is exploited in any hardware DNN implementation~\cite{moons2017envision,chen2016eyeriss}. Han, et al. (2015), use clustered training and trained pruning to reduce model sizes and propose a hardware accelerator optimized for their compression scheme~\cite{han2016eie}. Other recent hardware implementations propose solutions exploiting sparsity, either by speeding up~\cite{kim2017anovel} or by increasing energy-efficiency during sparse operation~\cite{chen2016eyeriss}~\cite{moons2017envision}. Some works~\cite{moons2017envision,han2016eie} expand upon that by also exploiting DNN's inherent tolerance to noise and variations by using reduced precision operators. This reduces arithmetic power consumption and compresses the models memory footprint, at the expense of a potential accuracy loss. All these works are implementations of existing neural networks, rather than cross-field optimizations.

\begin{figure*}[t]
     \centering
     \subfloat[][]{\includegraphics[width=0.3\textwidth]{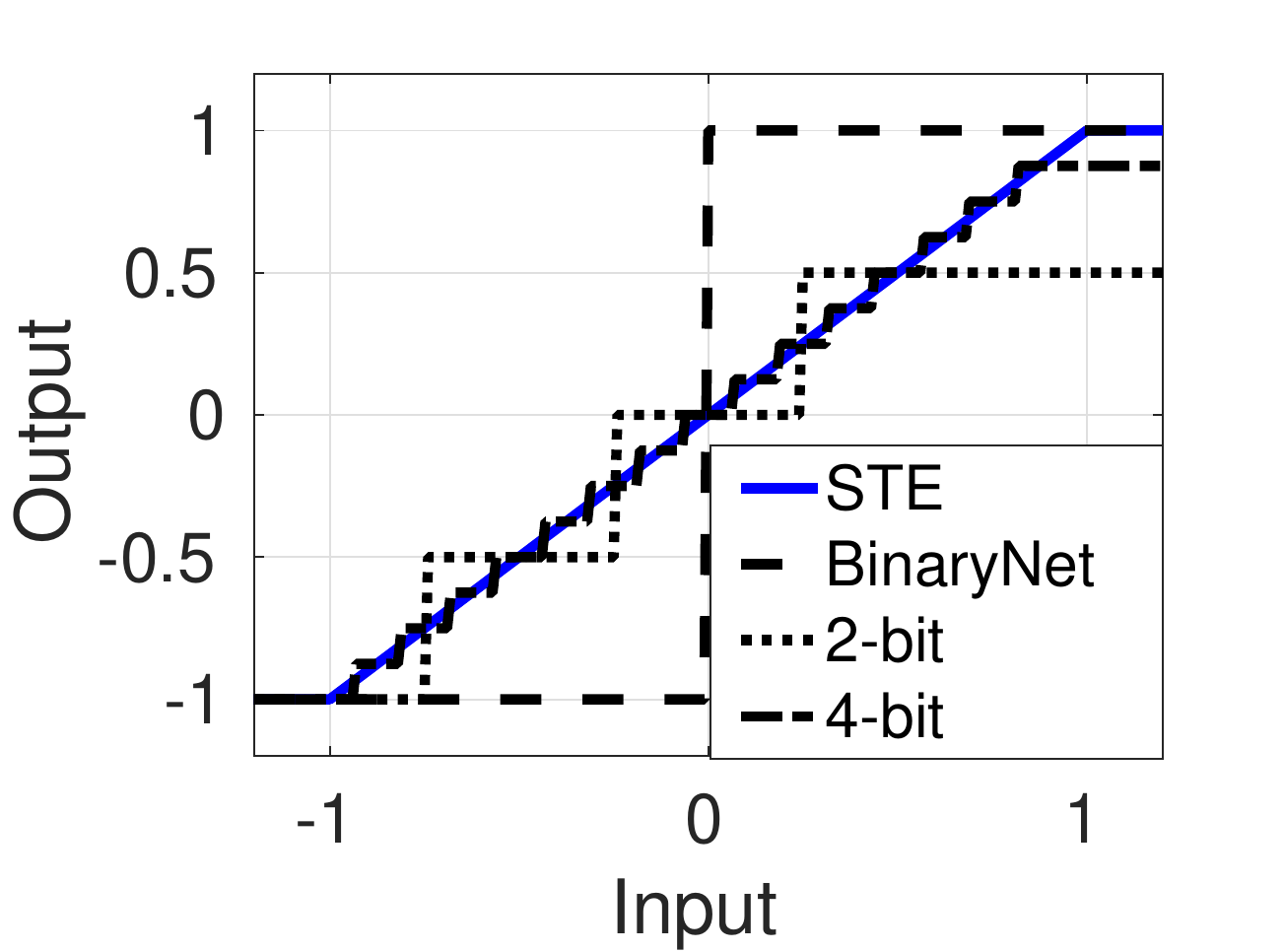}\label{fig:quantize_quantize}}\hfill
     \subfloat[][]{\includegraphics[width=0.3\textwidth]{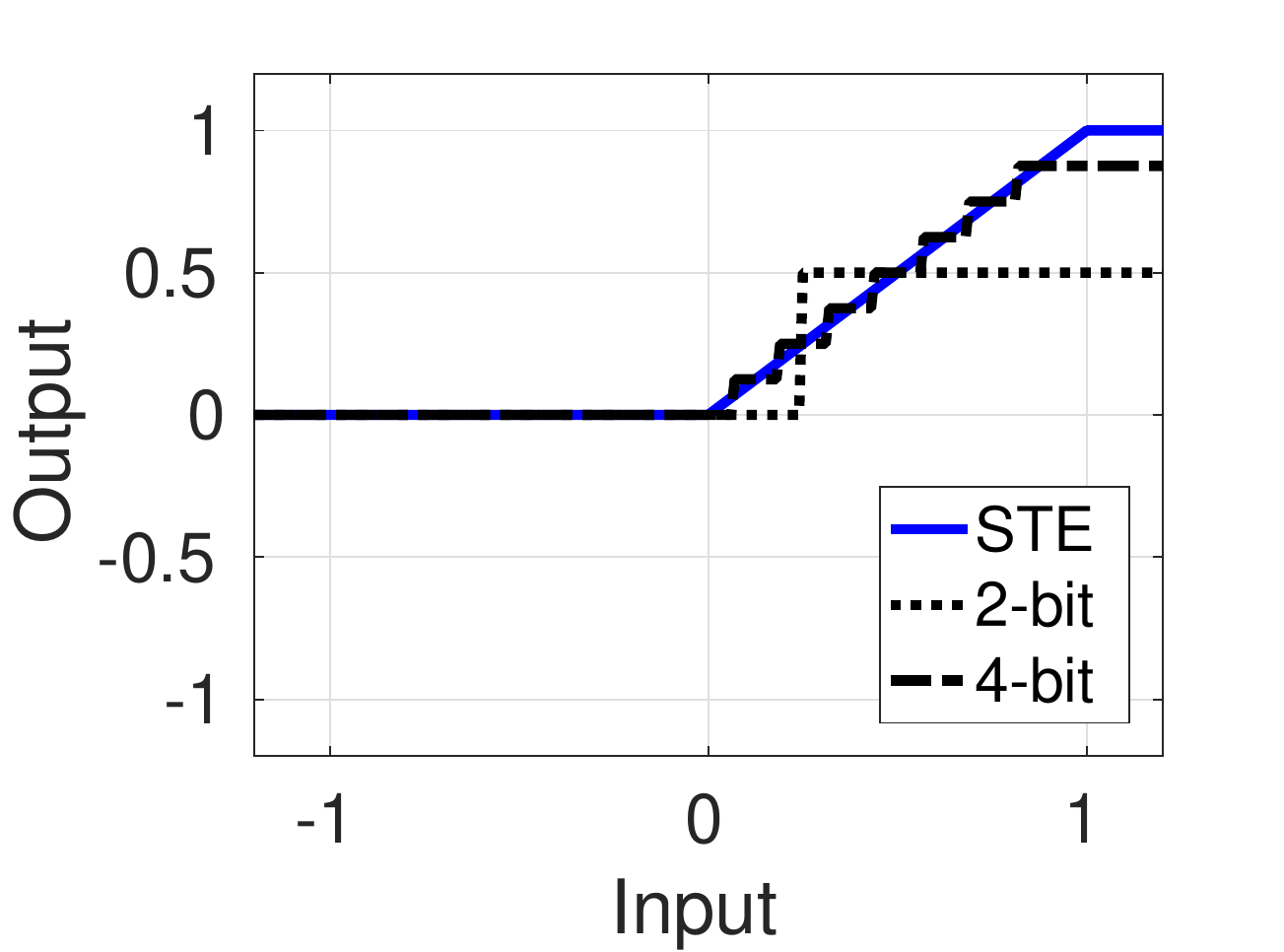}\label{fig:quantize_relu}}\hfill
     \subfloat[][]{\includegraphics[width=0.3\textwidth]{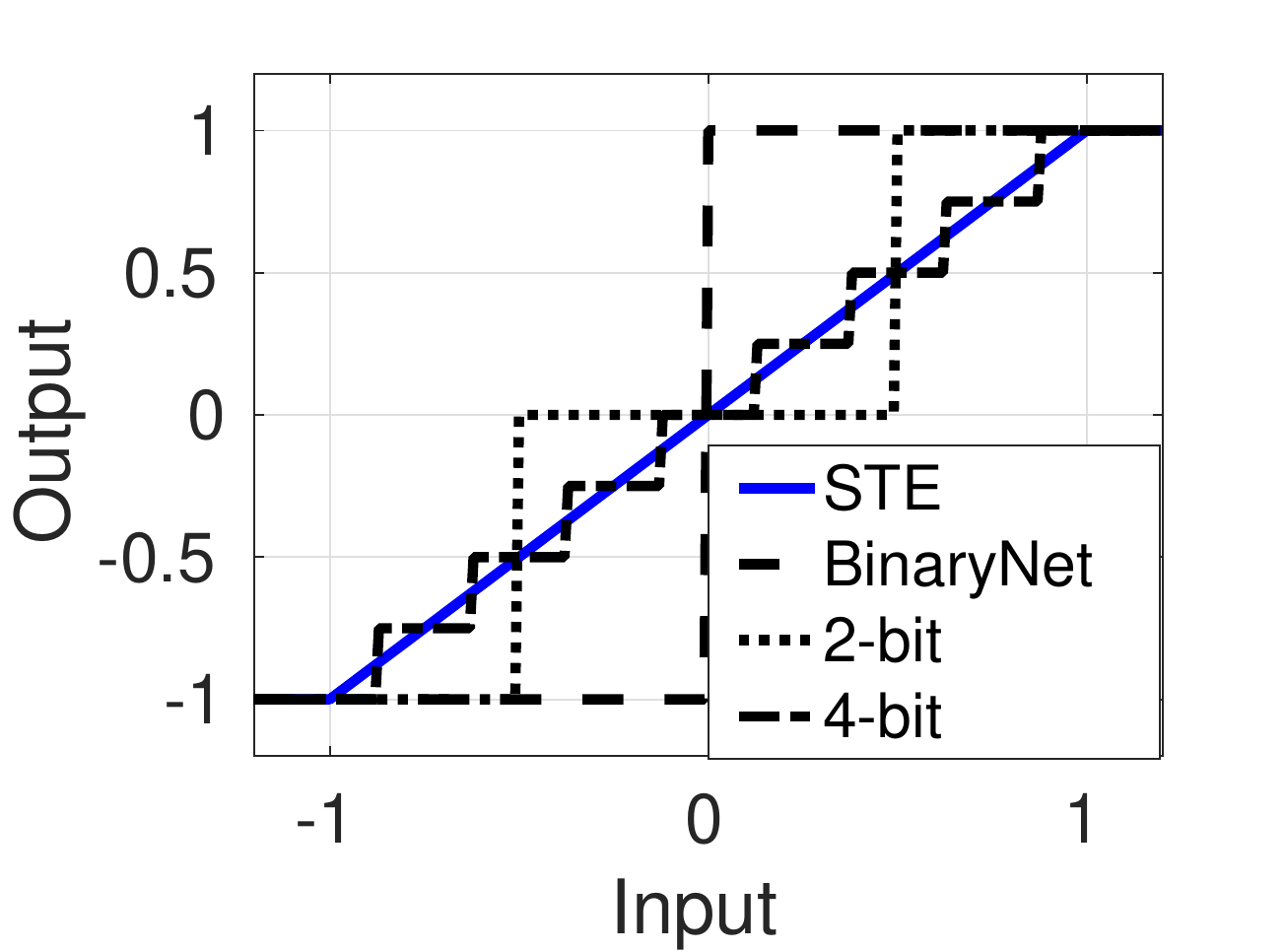}\label{fig:quantize_hardtanh}}
     \caption{(a) Weight quantization. (b) Quantized ReLU activation function. (c) Quantized hardtanh activation function. Straight-through estimators (STE) are used to estimate gradients. }
     \label{steady_state}
\end{figure*}

\textbf{Algorithmic innovations} towards new network architectures with a smaller memory and energy footprint have been made as well. Residual Neural Networks or ResNets~\cite{he2016deep} provide an alternative to VGG-type~\cite{simonyan2014very} networks, with considerably smaller network complexity and model sizes at iso-accuracy. Recent works have shown how computational complexity can be reduced by constraining the used computational precision during the training phase of a DNN. Most notably,~\cite{hubara2016binarized} can either constrain only the network weights, or both weights and activations to +1 and -1. This is particularly interesting from a hardware perspective, as such binary network topology allows replacing all costly multiply operations with an energy-efficient XNOR-operation. Other works have proposed ternarynets~\cite{zhu2016trained}, fixed-point analyses~\cite{moons2016energy} and fixed-point finetuning~\cite{gysel2016ristretto}. In~\cite{gysel2016ristretto}, the maximum achieved accuracy drops significantly at 2- or 4-bits.  All these techniques are ad-hoc, leading to sub-optimal results without offering full control over the used computational precision. Work presented in~\cite{pytorch}, shows a number of benchmarks can be quantized during inference down to 6-bits in the pytorch framework, but they do not retrain or train to an arbitrary number of bits and have no way to compare the energy consumption of the resulting modes. General Quantized Networks have been discussed in~\cite{hubara2016quantized} and~\cite{zhou2016dorefa}, where the authors run ad-hoc training-tests on specific network-topologies, but do not train them targeting minimum energy consumption.

Here, we offer explicit control over network quantization for any number of bits and any network topology through quantized training and link this to an inference energy model. This allows cross-optimizing both the used algorithm and the hardware architecture to enable always-on embedded applications. More specifically, our contributions are the following:
\begin{itemize}
\item We \textbf{generalize} BinaryNet training from 1- to Q-bits (BinaryNet to intQ), where Q can be any value $\in N$.
\item We \textbf{evaluate} the energy-accuracy-computational precision trade-off for QNN inference by linking network complexity and size to a full system energy-model.
\item We \textbf{conclude} energy consumption at iso-accuracy varies depending on the required accuracy, computational precision and the available on-chip memory. int4 implementations are often minimum energy solutions. 
\end{itemize}

\section{QNNs: Quantized Neural Networks}
\label{sec:quantized_neural_networks}
This section details our formulation of Quantized Neural Networks (QNN), which use only fixed-point representations for both weights and activations in training and inference. In essence, QNNs are the generalization of binary- and ternarynets~\cite{hubara2016binarized,zhu2016trained} to multiple bits, as in~\cite{hubara2016quantized,zhou2016dorefa}. Implementations in keras/tensorflow and lasagne/theano can be found on \textbf{https://github.com/BertMoons}.

In QNNs, all weights and activations are quantized to Q bits in a fixed point representation during training. All QNN models converge at all values of Q. The following quantization function is used to achieve this in the forward pass:
\begin{equation}
q = clip(\frac{round(2^{Q-1}\times w)}{2^{Q-1}},-1,1-2^{-(Q-1)})
\end{equation}
The Q=1 case is regarded as a special case, where $q=Sign(w)$, as in the original BinaryNet paper~\cite{hubara2016quantized}.
To successfully propagate gradients through discrete neurons, a "straight-through estimator" (STE) function is used for back propagation, which leads to fast training~\cite{bengio2013estimating}. If an estimator $g_q$ of the gradient $\frac{\partial C}{\partial q}$ has been obtained, the STE of $\frac{\partial C}{\partial w}$ is
\begin{equation}
g_w/g_q = hardtanh(w) = clip(w,-1,1)
\end{equation}
As in~\cite{hubara2016binarized}, all real valued weights are clipped during training onto the interval $[-1,1]$. Otherwise real-valued weights would grow large, without impacting the quantized weights. The weight quantization function $q(w)$ and STE are plotted in Fig.~\ref{fig:quantize_quantize} for different Q. Activations are done using either a quantized relu or hardtanh function.

\begin{figure*}[t]
     \centering
     \subfloat[][]{\includegraphics[width=0.5\textwidth]{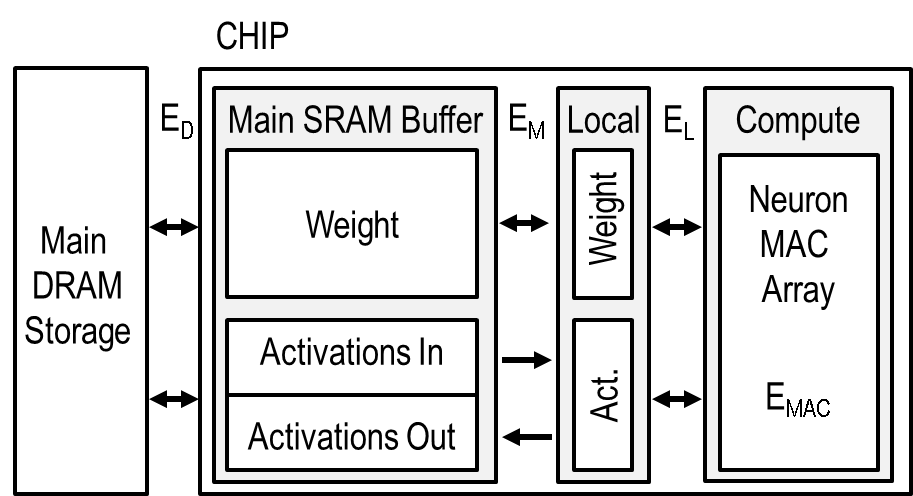}\label{fig:energy_model}}\hfill
     \subfloat[][]{\includegraphics[width=0.35\textwidth]{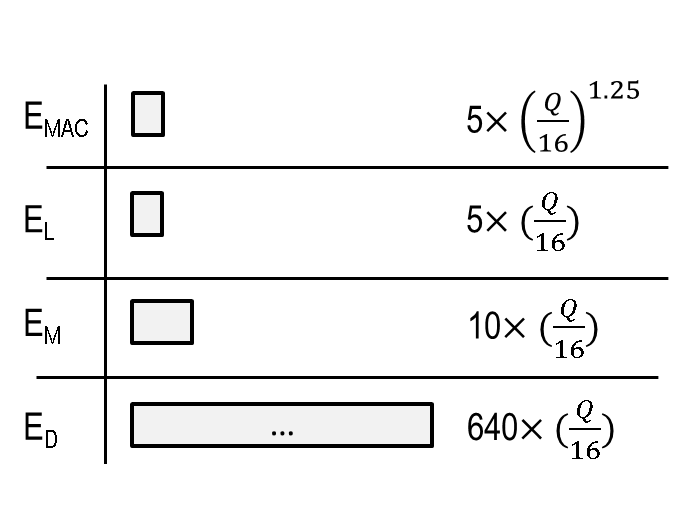}\label{fig:energy_models_values}}\hfill
     \caption{Used energy model for a NN platform based on~\cite{moons2017envision} and~\cite{Horowitz}. (a) High-level overview of the system architecture. (b) Relative energy consumption per equivalent Multiply-Accumulate (MAC) operation ($E_{MAC}$), read/write from the local ($E_{L}$) and main ($E_{M}$) SRAM buffers and per read/write from a large DRAM memory ($E_D$) of an intQ word.}
\label{fig:energy_modeling}
\end{figure*}



Multiple setups have been evaluated. Best results are achieved with the quantized ReLU function for int2, int4 and int8 and with the symmetrically quantized hardtanh function for the Q=1 case. As in~\cite{hubara2016binarized}, all real valued activations are clipped during training onto the interval $[-1,1]$. Every layer following an activation layer, will then have intQ inputs. The weight quantized ReLU- and hardtanh forward functions and STEs are plotted in Fig.~\ref{fig:quantize_relu} and~\ref{fig:quantize_hardtanh} for different Q.

In a QNN all the inputs to a layer are quantized to intQ, with the exception of the first layer, which typically has int8 pixels as its inputs. In a general case with M input bits where $M>Q$, an intQ layer can be performed as a sequence of $M/Q$ shifted and added dot products.

\section{Hardware energy model}
\label{sec:hardware_energy_model}

A generic, parameterized energy model, shown in Figure~\ref{fig:energy_modeling}, is used to assess the impact of QNNs on the energy consumption of a typical inference platform. Global energy per inference is the sum of the energy consumed in communication with an off-chip DRAM and the energy consumption of the processing platform itself. The total energy consumed per network inference is then:
\begin{equation}
\label{eq:total_mem}
 E_{inf} = E_{DRAM} + E_{HW}
\end{equation}

The sections below discuss a parameterized energy model, which can be customized to a wide variety of processing platforms by calibrating its parameters.

\subsection{Energy consumption of off-chip Memory-Access}
\label{subsec:energy_consumption_of_on-chip_memory-access}
The available memory in an always-on chip is inherently limited due to costs and leakage energy constraints and hence typically insufficient to store full models \textbf{and} feature maps. If this is the case, the chip will constantly have to communicate with a larger off-chip memory system. The cost of this interface is two orders of magnitude higher than a single equivalent MAC-operation~\cite{Horowitz}. Using less bits for weights and activations can hence be potentially more energy efficient, if the achieved network compression, both for weights and activations makes the network fit completely in on-chip memory. Off-chip DRAM access energy is modeled as:
\begin{equation}
\label{eq:dram}
 E_{DRAM} = E_{D} \times (s_{in}^2 \times c_{in} \times M/Q + 2\times f_{r} + w_{r})
\end{equation}
Where $E_{D}$ is the energy consumed per intQ DRAM access, as in Fig.~\ref{fig:energy_modeling}. $s_{in}$, $c_{in}$ and $M/Q$ are respectively the input image's dimensions, the number of input channels and the first-layer factor defined in section~\ref{sec:quantized_neural_networks}. $f_{r}$ and $w_{r}$ are the number of words that have to be re-fetched/stored from/to DRAM if a feature map or model does not fit in the on-chip memory.

\subsection{Hardware modeling}
\label{subsec:generic_asic_modelling}

The hardware platform, shown in Fig~\ref{fig:energy_modeling} is a typical processing platform for CNNs, based on~\cite{moons2017envision}. It contains a parallel neuron array, with a fixed area for $p$ MAC-units and two levels of memory. A large main buffer enables storing $M_W$-bits of weights and $M_A=M_{W}$-bits of activations, of which 50\% is used for the current layer's inputs and 50\% is used for the current layer's outputs. The small local SRAM or register file-buffers contain the currently used weights and activations. 
We model the relative energy consumption of SRAM-fetches and Multiply-Accumulate (MAC) operations according to Horowitz~\cite{Horowitz}. Here, the energy consumption of a read/write from/to the small local SRAM or Register file $E_{L}$ is modelled to be equal to the energy of a single MAC operation $E_{MAC}$, while accessing the main SRAM costs $E_{M}=2\times E_{MAC}$. Other operations, such as bias additions, quantized-ReLU and non-linear batch normalization are modeled to cost $E_{MAC}$ as well. All these numbers incorporate control-, data transfer and clocking overheads.
The total on-chip energy per inference $E_HW$ is then the sum of the compute energy $E_{C}$ and the cost of weight $E_{W}$ and activation accesses $E_{A}$:
\begin{align}
\begin{split} 
 &E_{C} = E_{MAC} \times ( N_{c} + 3\times A_s) \\ 
 &E_{W} = E_{M} \times N_{s} + E_{L} \times N_{c} / \sqrt p \\
 &E_{A} = 2 \times E_{M} \times A_{s} + E_{L} \times N_{c}/ \sqrt p 
 \end{split}
\end{align}

Here, $N_{c}$ is the network complexity in number of MAC-operations for partial product accumulation, $N_{s}$ is the model size in number of weights and biases and $A_{s}$ is the total number of activations throughout the whole network. Thus, $E_C$ is the sum of all energy consumed in partial product generation, biasing, batch-normalization and activation. The latter three are performed on all activations $A_s$, hence the term $3\times A_s$ in the $E_C$ equation. Weights are transferred once from the main to the local buffer and are then reused from there, leading to an equation for $E_W$. Here $\sqrt p$ is a reduction in memory energy, due to activation-level parallelism, as one weight is used simultaneously on $\sqrt p$ activations. A similar equation is derived for $E_A$, as activations are fetched/stored from/to the main buffer. The number of local activation fetches is reduced by $\sqrt p$ due to weight level parallelism, as one activation is simultaneously multiplied with $\sqrt p$ weights. The total level of parallelism $p$ is a function of Q, as the same area containing $p'$ 16-bit MACs, can hold $p'\times 16/Q$ intQ MACs. Similarly, an on-chip memory can store a variable number of weights, depending on the value of Q. A $2$Mb memory stores more then 2M weights, but only 131k 16-bit weights. If either the weight size or feature map size exceeds the available on-chip size, communication with a larger off-chip DRAM memory will be necessary, as discussed in section~\ref{subsec:energy_consumption_of_on-chip_memory-access}.

\section{Experiments}
\label{sec:experiments}

\begin{figure}
	\centering
     \subfloat[][]{\includegraphics[width=0.25\textwidth]{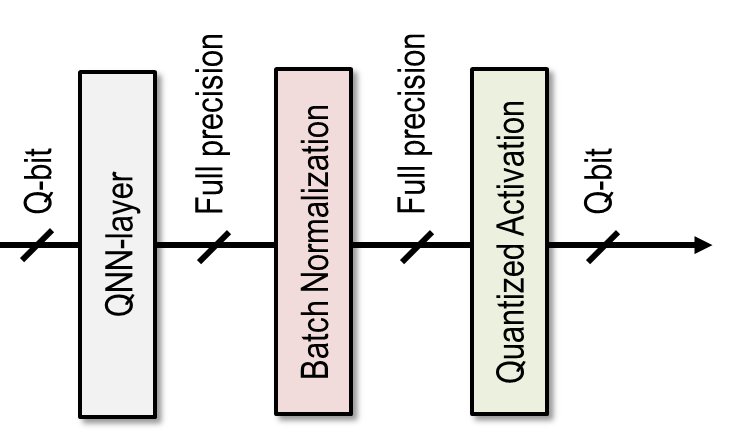}\label{fig:all_cnn_brick}} \hfill
     \caption{A QNN building block}
     \label{steady_state}
\end{figure}
\begin{table}[t]
\centering
\caption{Used QNN-topologies. $n_A,n_B,n_C,F_A,F_B,F_C$ and $n$ are taken as parameters. All used filters are $3\times 3$.}
\label{tab:topologies}
\begin{tabular}{l|c}
\textbf{Block} & \textbf{Classical QNN}  \\ \hline
Input & -  \\ \hline
Block A - $32\times 32$ & $n_A \times F_A \times 3\times 3$  + MaxPool(2,2) \\
Block B - $16\times 16$ & $n_B \times F_B \times 3\times 3$ + MaxPool(2,2)  \\
Block C~-~~~~~$8\times 8$ & $n_C \times F_C \times 3\times 3$  + MaxPool(2,2)  \\ \hline
Output & Dense - $4\times 4\times F_C$  \\
\end{tabular}
\end{table}
\begin{figure*}[!b]
     \centering
     \subfloat[][]{\includegraphics[width=0.3\textwidth]{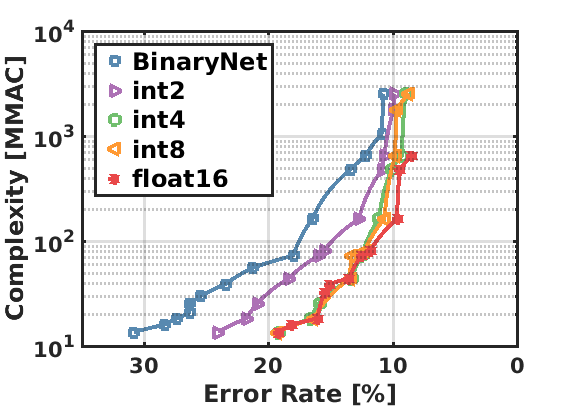}\label{fig:cifar10_complexity}}\hfill
     \subfloat[][]{\includegraphics[width=0.3\textwidth]{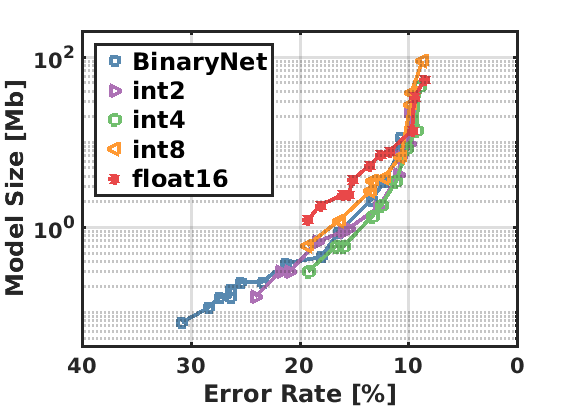}\label{fig:cifar10_model_size}}\hfill
     \subfloat[][]{\includegraphics[width=0.3\textwidth]{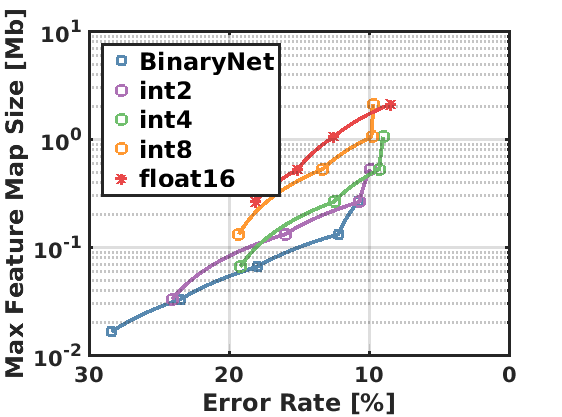}\label{fig:cifar10_feature_size}}
     \caption{ QNN networks on CIFAR-10~\cite{krizhevsky2009learning}. (a) computational complexity, (b) model size, (c) Maximum feature map size and the number of bits $Q$. }
     \label{fig:complexity_sizes}
\end{figure*}

\subsection{QNN topologies}
\label{subsec:qnn_topologies}

To quantify the energy-accuracy trade-off in QNNs, multiple network topologies are evaluated. This is necessary, as network performance not only varies with the used computational accuracy, but also with the network depth and width.

Each tested network contains 4 stages as shown in Fig.~\ref{fig:all_cnn_brick} and Table~\ref{tab:topologies}: 3 QNN-blocks, each followed by a max-pooling layer and 1 fully-connected classification stage as illustrated in Table~\ref{tab:topologies}. Each QNN-block is defined by 2 parameters: the number of basic building blocks $n$ and the layer width $F$. Every QNN-sequence is a cascade of a QNN-layer, followed by a batch-normalization layer and a quantized activation function, as shown in Fig.~\ref{fig:all_cnn_brick}. In this work $F_{Block}$ is varied from 32-512 and $n_{Block}$ from 1-3.

In order to reliably compare QNNs at iso-accuracy for different $n$, $n_{Block}, F_{Block}$ and $Q$, first the pareto-optimal floating-point architectures in the energy-accuracy space are derived . This can be done through an evolutionary architecture optimization~\cite{real2017large}, but here we apply a brute search method across the parameter space. Once this pareto-optimal front is found, the same network topologies are trained again from scratch, as QNNs with a varying number of bits.

\begin{figure*}[t]
     \centerline{
     \subfloat[][CIFAR-10~\cite{krizhevsky2009learning}]{\includegraphics[width=0.3\textwidth]{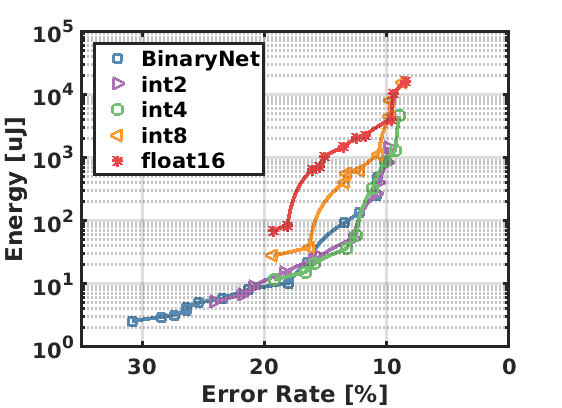}\label{fig:cifar10_energy}}\hfill
     \subfloat[][MNIST~\cite{lecun1998gradient}]{\includegraphics[width=0.3\textwidth]{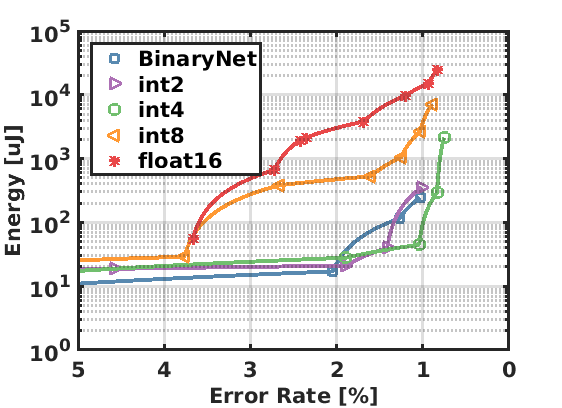}\label{fig:MNIST_energy}}\hfill
     \subfloat[][SVHN~\cite{netzer2011reading}]{\includegraphics[width=0.3\textwidth]{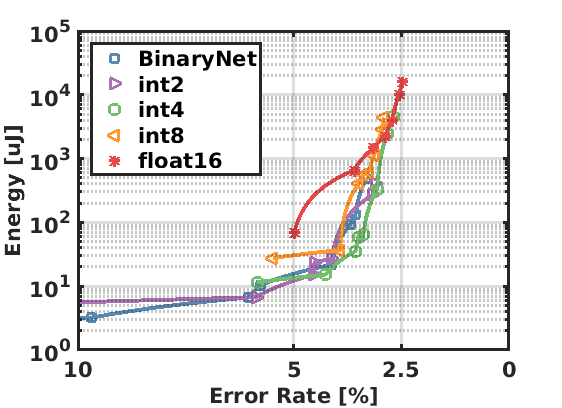}\label{fig:SVHN_energy}}}
     \caption{Error rate as a function of energy consumption for a typical $4$Mb chip. }
     \label{fig:energy_combo}
\end{figure*}

\begin{figure*}[t]
     \centering
     \subfloat[][1Mb on-chip MEM]{\includegraphics[width=0.3\textwidth]{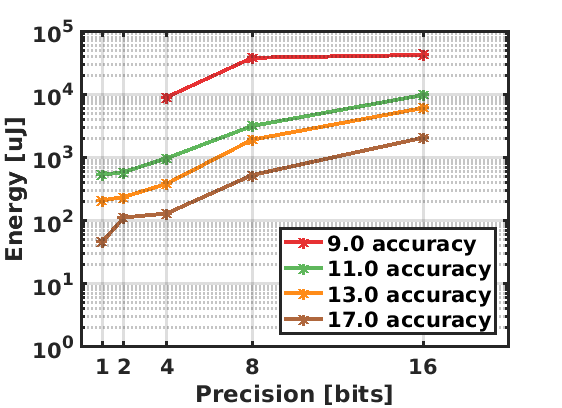}\label{fig:cifar10_minimum_energy_1Mbit}}\hfill
     \subfloat[][4Mb on-chip MEM]{\includegraphics[width=0.3\textwidth]{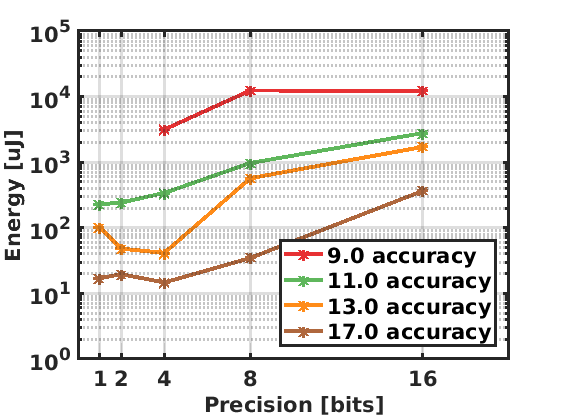}\label{fig:cifar10_minimum_energy}}\hfill
     \subfloat[][$\infty$Mb on-chip MEM]{\includegraphics[width=0.3\textwidth]{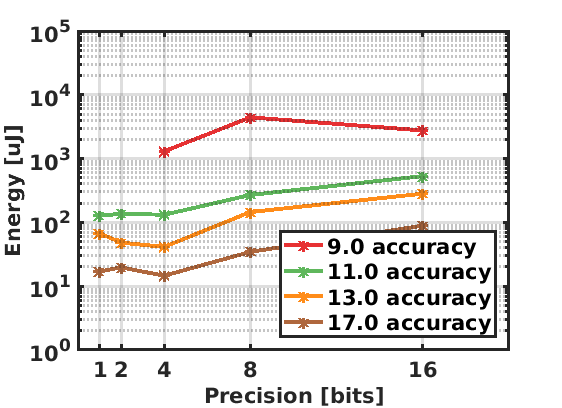}\label{fig:cifar10_minimum_energy_100Mbit}}
     \caption{Minimum energy plots for classical QNNs on CIFAR-10 for different chip-models.}
     \label{fig:minimum_energy_point_cifar10}
\end{figure*}

\subsection{Results and discussion}
\label{subsec:results_and_discussion}

The pareto-optimal set of QNNs is analyzed in search for a minimum energy network. In this analysis, we vary model parameters $M_W$ and $M_A$ and take $p=64\times (16)/Q$. 
Based on measurements in~\cite{moons2017envision}, we take $E_{MAC}=3.7pJ \times (16/Q)^{1.25}$. 

\textbf{Model sizes and inference complexity} are shown in Fig.~\ref{fig:complexity_sizes}. Here, computational complexity, model size and the maximum feature map size are compared as a function of error rate and $Q$ for the pareto optimal classical QNN set on CIFAR-10. Fig.~\ref{fig:cifar10_complexity} illustrates how the required computational complexity decreases at iso-accuracy if Q is varied from 1-to-16-bit, as networks with higher resolution require fewer and smaller neurons at the same accuracy. 
At $12\%$ error for example, the required complexity of a float16 network is $80$ MMAC-operations. Model complexity at iso-accuracy increases by $10\times$ to $800$ binary MMAC-operations. On the other hand, the model size in terms of absolute storage requirements increases with the used number of bits. This is illustrated in Fig.~\ref{fig:cifar10_model_size}. Here, an int4 implementation offers the minimum model size of only $2$Mb, at $12\%$ error rate. BinaryNets require 50\% more model storage, while the float16 net requires at least $4\times$ more. Fig.~\ref{fig:cifar10_feature_size} shows the storage required for feature maps as a function of network accuracy. If this size exceeds the available memory, DRAM access will be necessary. Here, BinaryNets offer a clear advantage over intQ alternatives. 

Fig.~\ref{fig:energy_combo} and Fig.~\ref{fig:minimum_energy_point_cifar10} illustrate the \textbf{energy consumption and the minimum energy point} for classical QNN architectures. Fig.~\ref{fig:energy_combo} shows the error-rate vs energy trade-off for different intQ implementations, for chips with a typical $4$Mb of on-chip memory. The optimal intQ mode varies with the required precision for all benchmarks. At high error rates, BinaryNets tend to be optimal. For medium and low error-rates mostly int4-nets are optimal. At an error-rate of $13\%$ on CIFAR-10 in Fig.~\ref{fig:cifar10_energy}, int4 offers a $>6\times$ advantage over int8 and a $2\times$ advantage over a BinaryNet. At $11\%$, BinaryNet is the most energy-efficient operating point and respectively $4\times$ and $12\times$ more energy-efficient than the int8 and float16 implementations. The same holds for networks with $10\%$ error. However, these networks come at a $3\times$ higher energy cost than the $11\%$ error rate networks, which illustrates the large energy costs of increased accuracy. In an int4 network run on a $4$Mb chip, energy increase $3\times$ going from $17\%$ to $13\%$, while it increases $20\times$ when going from $13\%$ down to $10\%$. Hence, orders of magnitude of energy consumption can be saved, if the image recognition pipeline can tolerate slightly less accurate QNN architectures.
Fig.~\ref{fig:minimum_energy_point_cifar10} compares the influence of the total on-chip memory size $M_W + M_A$. In Fig.~\ref{fig:cifar10_minimum_energy_1Mbit}, an implementation with limited on-chip memory, BinaryNets are the minimum energy solution for all accuracy-goals, as the costs of DRAM interfacing becomes dominant. In the typical case of $4$Mb, either BinaryNets, int2- or int4-networks are optimal depending on the required error rate. In a system with $\infty$Mb, hence without off-chip DRAM access, int2 and int4 are optimal. In all cases, int4 outperforms int8 by a factor of $2-5\times$, while the minimum energy point consumes $2-10\times$ less energy than the int8 implementations.

\section{Conclusion}
\label{sec:conclusion}
This work presents a methodology to minimize the energy consumption of embedded neural networks, by introducing QNNs, as well as a hardware energy model used for network topology selection. To this end, the BinaryNet training setup is generalized from 1-bit to Q-bit for intQ operators. This approach allows finding the minimum energy topology and deriving several trends. First, energy consumption varies by orders of magnitudes at iso-accuracy depending on the used number of bits. The optimal minimum energy point at iso-accuracy varies between 1- and 4-bit for all tested benchmarks depending on the available on-chip memory and the required accuracy. In general, int4 networks outperform int8 implementations by up to $2-6\times$. This suggests, the native float32/float16/int8 support in both low-power always on applications and high performance computing, should be expanded with int4 to enable minimum energy inference.







\small
\bibliographystyle{unsrt}
\let\oldbibliography\thebibliography
\renewcommand{\thebibliography}[1]{%
  \oldbibliography{#1}%
  \setlength{\itemsep}{4.5pt}%
}
\bibliography{references}

\end{document}